\documentclass[letterpaper, 10 pt, conference]{ieeeconf}  

\IEEEoverridecommandlockouts                              

\newcommand{\RNum}[1]{\uppercase\expandafter{\romannumeral #1\relax}}

\usepackage{graphics} 
\usepackage{graphicx}
\graphicspath{{./figure/}}
\usepackage{subcaption}
\usepackage[font=small]{caption}

\usepackage{bm}

\usepackage[noadjust]{cite}

\usepackage{hyperref}
\usepackage{amsmath} 
\usepackage{amssymb}  
\usepackage{algorithm}
\usepackage{algorithmic}

\usepackage{booktabs}
\usepackage{multirow}
\usepackage{cleveref}
\usepackage{color}

\title{\LARGE \bf
	FingerVision Tactile Sensor Design and Slip Detection Using Convolutional LSTM Network
}

\author{Yazhan Zhang$^{1}$, Zicheng Kan$^{1}$, Yu Alexander Tse$^{1}$, Yang Yang$^{1}$, \textit{Member, IEEE} \\
	and Michael Yu Wang$^{2}$, \textit{Fellow, IEEE}
	\thanks{*Research is supported by the Hong Kong Innovation and Technology
Fund (ITF) ITS-018-17FP.}
	\thanks{$^{1}$Y. Zhang, Z. Kan, Y. Tse and Y. Yang are with the Department of
Mechanical and Aerospace Engineering, Hong Kong University of Science
and Technology, Hong Kong (e-mail: yzhangfr@connect.ust.hk; zkan@connect.ust.hk; yatse@connect.ust.hk; rayang@ust.hk).}%
	\thanks{$^{2}$M. Y. Wang (corresponding author) is with the Department of Mechanical
and Aerospace Engineering and the Department of Electronic and Computer
Engineering, Hong Kong University of Science and Technology, Hong Kong
(tel.: +852-34692544; e-mail: mywang@ust.hk).}%
}

\begin{document}

\maketitle
\thispagestyle{empty}
\pagestyle{empty}

\begin{abstract}

Tactile sensing is essential to the human perception system, so as to robot. In this paper, we develop a novel optical-based tactile sensor ``FingerVision'' with effective signal processing algorithms. This sensor is composed of soft skin with embedded marker array bonded to rigid frame, and a web camera with a fisheye lens. While being excited with contact force, the camera tracks the movements of markers and deformation field is obtained. Compared to existing tactile sensors, our sensor features compact footprint, high resolution, and ease of fabrication. Besides, utilizing the deformation field estimation, we propose a slip classification framework based on convolution Long Short Term Memory (convolutional LSTM) networks. The data collection process takes advantage of the human sense of slip, during which human hand holds 12 daily objects, interacts with sensor skin and labels data with a slip or non-slip identity based on human feeling of slip. Our slip classification framework performs high accuracy of 97.62\% on the test dataset. It is expected to be capable of enhancing the stability of robot grasping significantly, leading to better contact force control, finer object interaction and more active sensing manipulation.

\end{abstract}


\section{INTRODUCTION}
Tactile sensing plays a central role in human haptic perception system and during prehensile manipulation. In \cite{johansson1979tactile}, G. Westling et al. experimented with volunteers on tactile sensation to find that without tactile feedback, human have great difficulty in maintaining stable grasping. To execute complex tasks in dynamic environments, robots are expected to be equipped with perception capability similar to human level. Robotic tactile sensor is thus a critical component for adaptive robot system. Besides, effective encoding of contact interaction events between robot and environment directly affects success or failure of various manipulation tasks, among which slip detection is essential when trying to secure stable grasping and performing dexterous manipulation. In this paper, our problem is two-folded, one is to develop high performance optical-based tactile sensor, another is to propose a slip detection framework based on the sensor that we have developed.

\subsection{Optical-based tactile sensor}
These decades have witnessed large development in robotic tactile sensors.
In \cite{Dahiya:2010:TSH:1771964.1771965}, comprehensive review of robot tactile sensing is given. Although researchers have been devoted to investigating robot tactile sensors and some of them are already available commercially, e.g., BioTAC in \cite{wettels2009multi}, a usable out-of-shelf, high resolution, high dynamic and stable robotic tactile sensor is still yet to come.

Recently, optical-based tactile sensors have been attracting more and more attention. Compared to tactile using other transducing methods, optical tactile sensors can output high resolution and
sensitivity signal with a relatively large working area by converting contact signals into image. In principle, optical sensor
acts as a mapping interface between contact surface deformation and contact information including shape, contact
wrench, and higher level occurrence of slip, etc. 
In \cite{ohka1996data}, M. Ohka et al. present a tactile sensor
made of rubber and pyramid-shape acrylic with a CCD camera to capture the deformation of rubber, which can detect three axial forces. 
Nicola J. Ferrier et al. \cite{ferrier2000reconstructing} present their design of tactile sensor composed of a deformable
membrane and a camera, and meanwhile propose reconstruction algorithm for elastic contact surface. In that work,
a coarse three-dimensional shape of contact object is reconstructed by tracking of markers movement within contact
surface. 
K. Sato et al. \cite{sato2010finger} fabricate and finger-shape tactile sensor ``GelForce'' with embedded markers
to obtain force vectors field. 
A. Yamaguchi et al. \cite{yamaguchi2016combining} implement a sensor called ``FingerVision'' (name firstly
used in \cite{yamaguchi2017implementing}) with markers embedded in elastic
layer and captured by camera. This elastic layer of the sensor is transparent, which makes it capable of capturing
both marker movements and external object movement that is a desirable property for target approximating detection. 
In \cite{johnson2009retrographic}, tactile sensor ``Gelsight'' was first published, which is reported to be capable to capture high resolution reconstruction of contact geometry. This sensor is lately
updated to be of standardized fabrication, high repeatability, and great reconstruction precision \cite{dong2017improved}. 
Yuan et al.
\cite{yuan2015measurement} print markers on Gelsight surface to endow it with a sense of physical information including normal, tangential,
torsional forces.

Inspired by these previous works, we devise an optical-based tactile sensor that is able to capture movement of object contacted with sensor skin. This novel system, enhances the performance stability, durability and resolution of tactile sensor, decreases the difficulties in fabrication procedure as well. The sensor is mainly composed of a web camera (640$\times$480 pixels) as transduction interface and markers embedded soft layer (with markers grid spacing of 1.5mm) imitating human skin with  distributed receptors. Our sensor is capable of encoding tangential wrench, which is vital to characterize interaction between sensor and objects, e.g., contact making and breaking, slip and incipient slip, etc. Further implementation of slip detection application making use of these encoded contact information is demonstrated.

\subsection{Slip detection}
Human not only grasp objects dexterously with sensitive feedback of slip, but also make use of slip to do fine manipulation, for instance, rolling pen between fingers. This sensation input for robot system to execute manipulation tasks is also important. Slip occurs when contact force is insufficient, which usually required the robot to automatically adjust grasping plan accordingly.

Owing to the wide range of potential applications of slip detection in the robot system, people have been developing slip detection methods based on various types of sensors, including force transducers \cite{yamada1994tactile}, accelerometers \cite{howe1989sensing}, capacity tactile array \cite{heyneman2016slip} and optical-based tactile sensors \cite{yuan2015measurement} in the past. M. T. Francomano et al. \cite{francomano2013artificial} present a thorough review of slip detection methods. Most of slip detection methods fall in the group of algorithms including signal spectral analyses \cite{heyneman2016slip}, optical flow algorithms \cite{alcazar2012estimating}, 
optical displacement field measurements \cite{ikeda2004grip} to predict slip. Using Gelsight sensor with markers embedded, Yuan et al. \cite{yuan2015measurement} and Dong et al. \cite{dong2017improved} detect slip occurrence by measuring markers displacement field distribution entropy and the relative displacement between object texture and markers respectively. However, the performance of their detection algorithms varies with different contact objects and contact conditions, which makes them inconsistent while dealing with large set of target objects.

Neural networks have been shown effective on challenging problems and to generalize well on wide range of targets with proper treatments. Slip detection schemes that utilize recurrent neural networks have been reported to be successful since slip is an inherently temporal behavior during contact. In \cite{li2018slip}, J. Li et al. propose a convolution neural network (CNN) combined with LSTM, taking both Gelsight image and external monitoring image of contact as input to detect slip occurrence, and achieves an accuracy of 88.03\% on relatively small size dataset. In \cite{van2018slip}, they performs dimension reduction and spectral analysis of signals from multiple commercial tactile sensors before being fed into LSTM network and generate high quality slip detection with accuracy around 90\%.

Encouraged by these success,  we propose a Convolutional LSTM \cite{shi1506convolutional} slip detection framework with  the ``FingerVision'' tactile sensor we develop. The convolutional LSTM network captures spatiotemporal correlations better than the concatenation of convolutional neural network (CNN) and LSTM model because the former model encodes spatial information simultaneously with temporal transition, which is suitable to our problem considering our data being image frames. We show that our structure outperforms similar methods \cite{li2018slip} in terms of slip/non-slip classification accuracy, computation cost and transferability with fabrication variance.

The rest of this paper is organised as follows: Section \RNum{2} explains the implementation of ``FingerVision'' which includes
conceptual design, fabrication process and algorithm pipeline. Section \RNum{3} demonstrates the feature tracking outcomes and
deformation field results. Section \RNum{4} illustrates the framework and dataset collection used for slip-detection and
a demonstration of the slip detection performance is given. Finally, in Section \RNum{5}, conclusion is drawn and future work is discussed.

\section{FingerVision Design and Fabrication}

This section introduces the conceptual design of FingerVision, followed by fabrication process in detail.

\subsection{Conceptual Design}

The basic scheme to get contact information from optical-based tactile sensor is by inducing large
deformation of material in contact region for camera to capture the deformation. Similar to \cite{yamaguchi2016combining}, our sensor consists of elastomer with embedded markers, camera and supporting frame.  Although the tactile sensor developed in \cite{yamaguchi2016combining} is versatile on sensing multiple modalities, the relatively low stability during real robotic action makes
their sensor not an ideal tactile component for robots. In the work of \cite{dong2017improved}, Gelsight sensor is compact,
easy to use, able to reconstruct super fine contact surface, however, extra calibration of correspondence between color and surface normal is required. 
Our sensor design emphasizes on stability, durability, readiness to use and adopts optical-based sensor scheme for its superior properties. 

\begin{figure}
	\centering
	\includegraphics[width=0.46\textwidth]{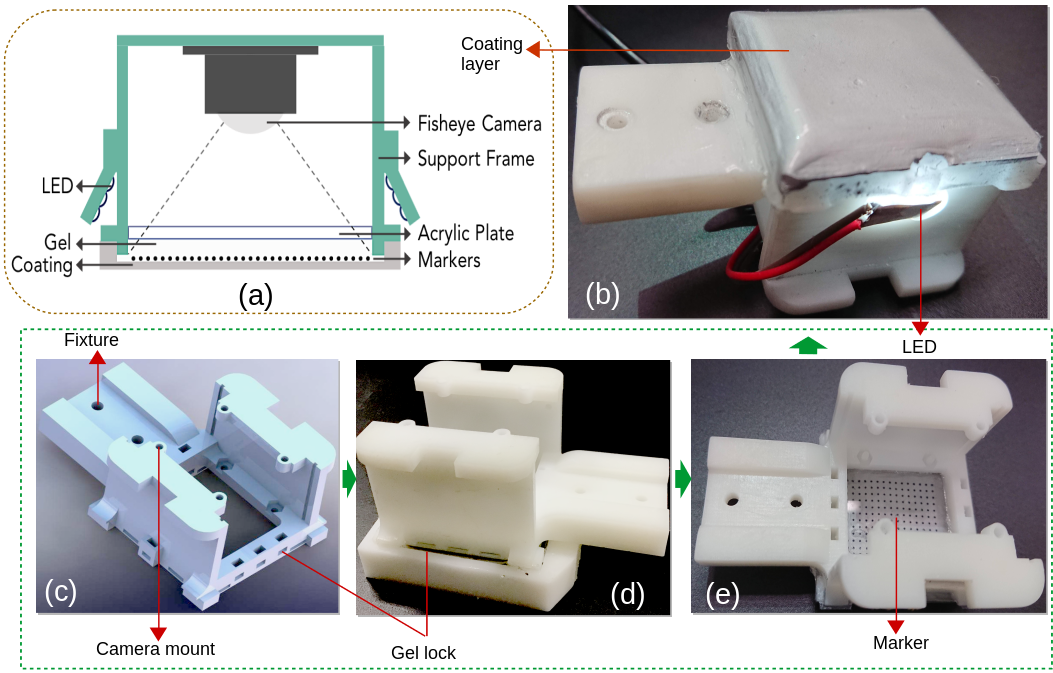}
	\caption{Design and fabrication of FingerVision tactile sensor. (a). Conceptual diagram of sensor. (b). Finish of tactile sensor. (c). Fabrication process. (c). Rendered model of sensor supporting frame. (d). Casting in action with frame in mold. (e). Intermediate step after markers attached.}
\end{figure}

For the design of optical-based tactile sensor, the following issues need to be taken into account. 
\begin{itemize}
	\item \textit{Functionality}. To enhance the stability of markers tracking, consistent background and even illumination are critical. Wide view angle fisheye is necessary because large view field is needed given short focusing range.
	\item \textit{Number
of components}. Large number of components increases degree of difficulty during fabrication process and augments the complexity of system, which requires more characteristics to be set properly.
	\item \textit{Compactness}. Compactness of sensor determines whether it can be installed on robot fingers and limits the range of tasks it can be applied to.
\end{itemize}
 
Conceptual design of FingerVision is depicted in Fig. 1(a). Fisheye camera, a transparent elastic
layer, and the supporting frame comprise the hardware of the sensor. Inside the elastomer,
markers are used as tracking feature points reflecting the displacement of material in
neighborhood, and with a compact 2D marker array, dense displacement field of contact
surface can be acquired. We choose fisheye as transducer due to the relative larger field of view,
which achieves a better trade-off between the compactness of sensor and large sensing area. 
High tracking stability is guaranteed with even illumination, rather than the way of light guiding through acrylic plate as reported in \cite{yuan2015measurement}, in our design, the LED light source
is put outside of white supporting frame (via 3D printing), and the light transmits through the resin
with large diffusion , which consequently illuminates the sensing area evener. Unlike the design in \cite{yamaguchi2016combining}, we coat the gel layer with another reflective layer to block external light sources that can deteriorate the tracking
performance of sensor. Similarly, the supporting frame is designed to be fully enclosed to increase tracking stability. 

Fig. 1(c) and (d) show the rendered 3D model of sensor frame and mold for the casting
of Gel layer respectively. After iterations of design, fabrication and redesign, it is found that
distributed locking holes in the outer ridge of frame are also important to maintain firm attachment between silicone rubber layer and the frame, learned from \cite{yamaguchi2017implementing}.

\subsection{Fabrication}

Our fabrication process includes three steps as shown below, in addition, material and components are also described.

1) \textit{3D printing of sensor frame and casting mold}. Benefiting from the easily available 3D printing, we are able to accelerate the prototyping process. The sensor frame and
mold are made of Epoxy Resin with SLA 3D printing technology,
possessing properties of smooth surface finish, good accuracy, and mechanical strength. 

2) \textit{Elastomer casting and marker embedding}. As for material of the elastomer, we
choose a transparent silicone rubber \textit{Solaris} from Smooth-On, Inc. This material, come
with two parts, A and B, and cures with weight ratio of \textit{1A:1B}. The casting lasts for 24 hours in room temperature and can be accelerated in the oven, the shore hardness of cured
silicone rubber is 15A. To put markers on the surface, the markers are patterned on a water transfer paper
first, and follow the product instruction to transfer ink from paper to gel surface. In this manner,
the markers density is much higher than that of the method used in \cite{yamaguchi2016combining} by
embedding plastic beads into the elastic layer. Fig. 1(d) shows the casting of sensor prototype in action.

3) \textit{Assembly of sensor components}. The camera we use is a commercially available CCD web
camera with fisheye lens. Before the camera is installed on the frame,
fisheye camera calibration is required. Here we adopt chessboard camera calibration method for
fisheye lens to estimate intrinsic matrix. Since the working focus range of camera is short compared to other application scenarios, calibration chessboard needs to be printed with small size as well.
After these three steps, we obtain the optical-based tactile sensor as shown in Fig. 1(b).

\section{Sensor Signal Processing}

\begin{figure}
	\centering
	\includegraphics[width=0.48\textwidth]{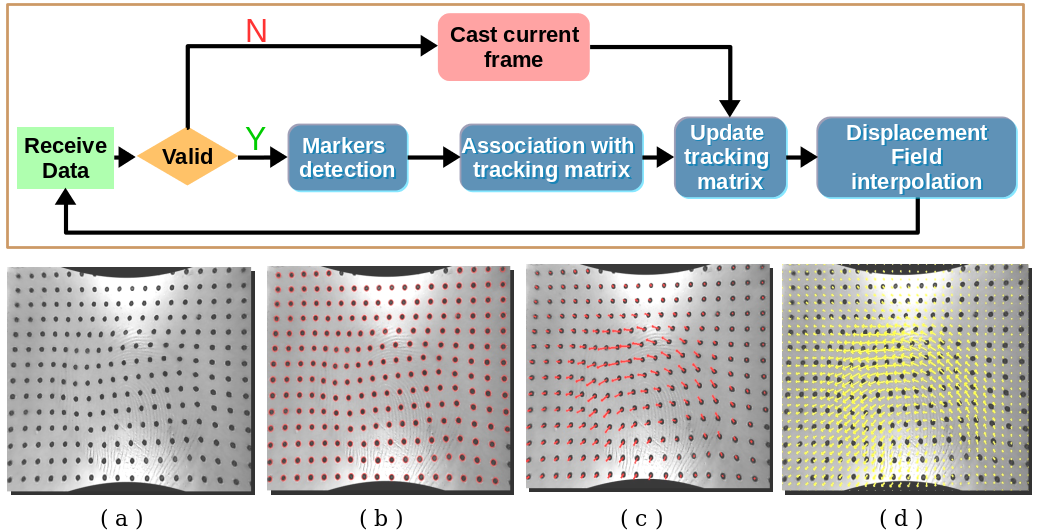}
	\caption{Sensor data processing pipeline and corresponding results (with sensor pressed by thumb). (a). Captured input image. (b). Key points detection result. (c). Marker tracking vectors. (d). Interpolated marker displacement field.}
\end{figure}

After acquiring images of contact region, additional processing is conducted to obtain contact deformation field. Fig. 2 shows the signal acquisition and processing pipeline with results step-by-step. Markers detection,
tracking algorithm and deformation field interpolation are three essential components of
the system. For the markers are 2D array of black dots on a lighter background, we adopt
blob detection to calculate centers of every black dots. With fully enclosed chamber and even illumination, the blob detection works stably with fine-tuned parameters. A sample image and its key points are shown in
Fig. 2(a) and (b). Compared with transparent sensing elastomer design in \cite{yamaguchi2016combining}, our sensor outperforms in terms of robustness in various working scene.

\begin{algorithm}
	
	\caption{Markers Tracking Algorithm}
	
	\begin{algorithmic}
		
		\WHILE {TRUE}
		\STATE Capture image
		\STATE $centroids \gets blobDetection(image)$
		\IF {First loop}
		\STATE $T, initPos \gets centroids$ 
		\STATE Continue
		
		\ELSE 
		\STATE $C \gets KNN(initPos, centroids)$
		
		\FOR {$k \gets 1$ \TO  $N$}
		\STATE $lastPoint \gets T_k$
		\STATE $currentPoint \gets C(lastPoint)$
		
		\IF {$dist(lastPoint, currentPoint) \leq d$}
		\STATE Update lastPoint in T with currentPoint
		\ENDIF
		\ENDFOR
		\ENDIF

		\ENDWHILE
	\end{algorithmic}
\end{algorithm}

Recovering of deformation field requires tracking of markers sequentially on time domain. After the centroids of markers are obtained in marker detection module, we proceed to
track the movement of every marker individually. Commonly there are two approaches for tracking problems, one is by updating correspondences between consecutive frames and then tracing back to the initial frame to obtain displacements, another is by registering correspondences of markers between first frame and current frame. For registration methodology, the state-of-the-art of non-rigid registration methods are
coherent points drift method ( CPD) \cite{myronenko2007non} and elastic Iterative Closet Points (ICP) \cite{stewart2003dual} method. Although registration methods is able to compute correspondences between arbitrary two frames, which makes it more robust and less interdependent between frame to frame, the iterative optimization scheme severely deteriorates the real-time sensing capability.

Therefore, we adopt the scheme of updating correspondences frame by frame, and the tracking algorithm with
tolerance to markers detection failure is developed. Firstly K-nearest neighborhood
method is used to obtain nearest points correspondences $C$ between previous frame and
current frame, and then corresponded pairs with euclidean distance larger than manually set threshold $d$ are rejected, which is reasonable by assuming that the displacements of points are small
with the short time elapsed between consecutive frames. Thereafter, valid point
correspondences are updated in tracking matrix $T$ as latest correspondence matrix
associating initial marker positions and that of last updated markers.

Deformation of elastic material is naturally smooth with its low pass filter effect for external excitation. 
Therefore, to further increase the density of deformation field and obtain
vector field with fixed size in fixed locations to facilitate further usages, we perform displacement
field interpolation based on the tracked displacement vectors. Here radius basis function (RBF) interpolation
method is selected to obtain smooth interpolation with relatively lower computation cost. The interpolation result is shown in Fig. 2. The frame rate of our set up (Intel Core i7-7700K@4.2GHz, 8 cores) with the processing algorithm is 15 FPS, which is adequate for various robotic tasks.

\begin{figure}
	\centering
	\includegraphics[width=0.45\textwidth]{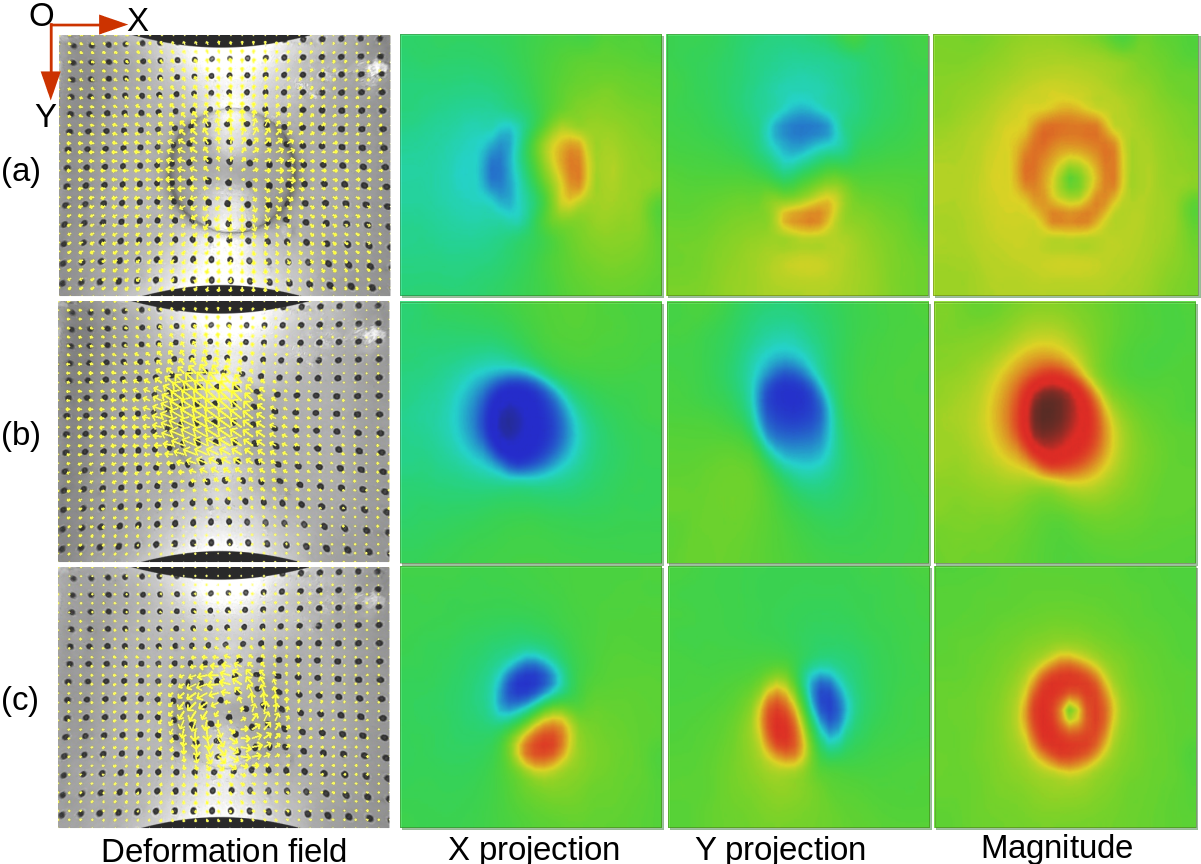}
	\caption{Deformation field and orthogonal projection when excited with different directional forces (Colormap to value: Green: neutral, Red: positive, Blue: negative). (a). Applied with Normal force. (b). Applied with normal and tangential force. (c). Applied with normal and rotational force along contact surface normal.}
\end{figure}

Contact motion can be decomposed into normal pressing, tangential translation,
torsional motion. However, in real life, contact breaking happens when pure tangential force is applied without normal pressing. Fig. 3 shows the deformation field obtained with FingerVision sensor while being excited by pure normal, tangential coupled with normal, and torsional coupled with normal forces with a doom shape indent head. It is obvious that our sensor signal can distinguish different contact forces successfully, which is useful for various robot manipulation tasks. Application to slip detection is implemented in next section and it is shown to be effective.

\section{Slip Detection}
In this section, we propose a learning based slip detection framework making use of contact deformation field generated by FingerVision sensor. The framework is demonstrated to accurately classify contact into slip and non-slip states using tactile information only.
\subsection{Model Description}

\begin{figure*}
	\centering
	\includegraphics[width=0.7\textwidth, height=3.5cm]{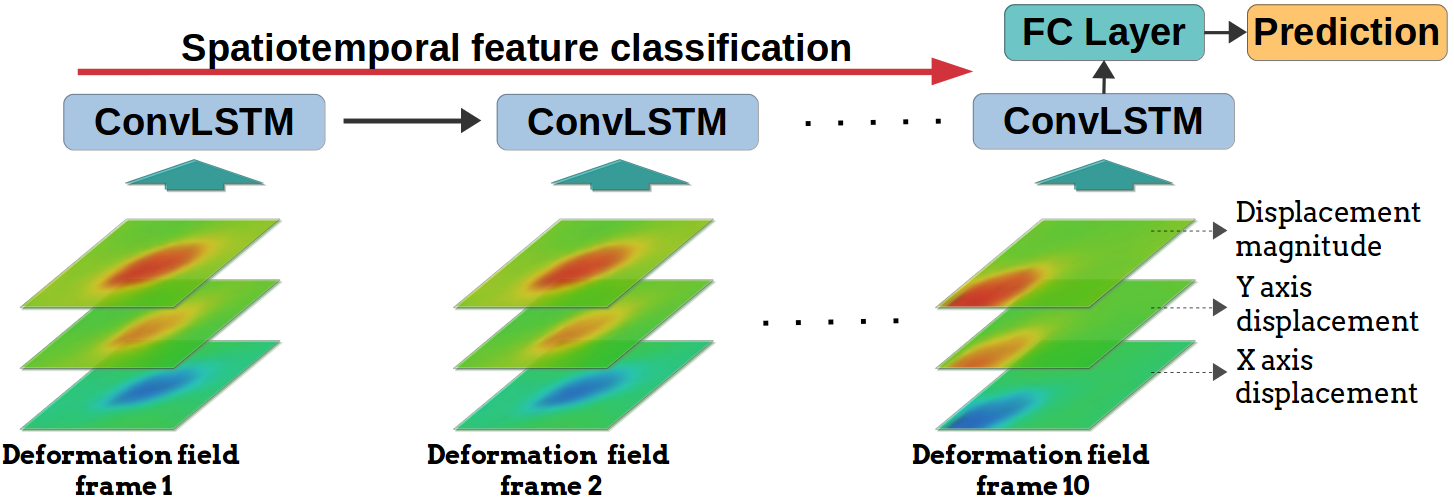}
	\caption{The architecture of slip detection framework using convolutional LSTM network and FingerVision Data.}
\end{figure*}

The architecture of slip detection framework is shown in Fig. 4. Knowing that slip occurrence is a spatiotemporal event on the contact surface, we adopt convolutional LSTM for its capability of capturing spatiotemporal features transition simultaneously \cite{shi1506convolutional}. Compared to the structure concatenating time distributed CNN with LSTM as \cite{li2018slip}, convolutional LSTM network is preferred because spatial feature extraction parameters (convolution kernel weights) are also the hidden node parameters and back propagation update is performed on both domains simultaneously. Besides, the well-performing CNN used in \cite{li2018slip} Inception V3 model contains over 7 million parameters \cite{canziani2016analysis}, which makes the forward pass computation too demanding for real-time implementation. In this perspective, It is alleviating that our model has drastically lower 269,826 parameters. We feed 10 frames of multi-channel images containing deformation field with x axis projection, y axis projection, and magnitude value jointly into convolutional LSTM network sequentially. Since the multiplication operation in common LSTM is replaced by convolutional operation \cite{shi1506convolutional}, the output of LSTM is also 3D tensors. The output of convolutional LSTM is flatten and fed to fully connected layer (FC layer) afterwards. LSTM layer with 64 hidden nodes with $5\times5$ kernal size is set for our model. Multiple layer LSTM structure is beneficial for long term dependency learning, however, considering our data is collected in a relatively short period, single layer structure is constructed. Finally, the FC layer outputs two variables representing the probability of stable contact or slip.

As for the training setup, Pytorch package is used to build the network and our model is trained on Nvidia GeForce 1060 (6GB graphic memory). In the training phase, The network weights are initialized with random small values and we choose two classes cross entropy as loss function. Adam optimizer with learning rate of $1\times10^{-5}$ is used for network parameters update.

\subsection{Experimental Analysis}

\begin{figure}
	\centering
	\includegraphics[width=0.45\textwidth]{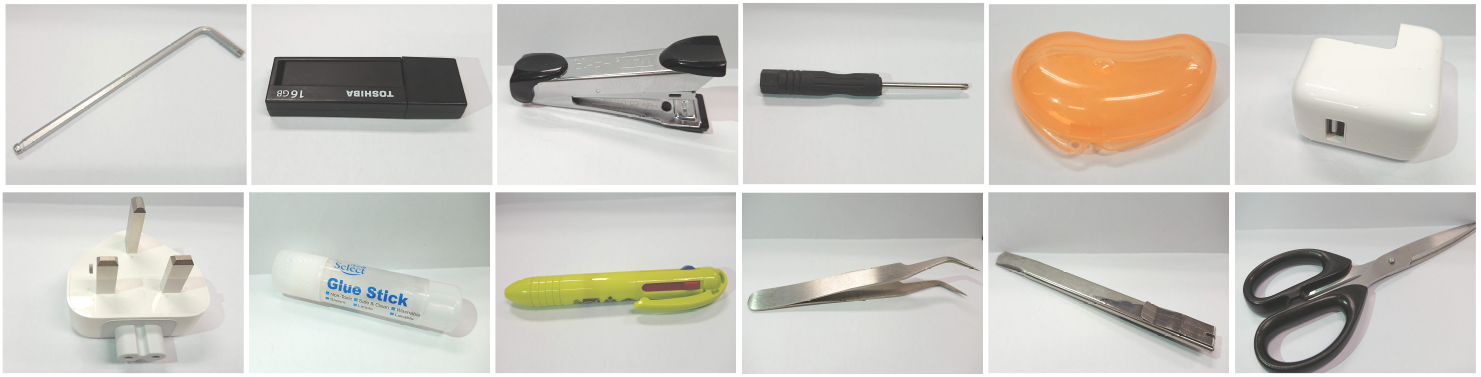}
	\caption{12 daily objects used in dataset collection.}
\end{figure}

\textbf{Data collection}. To distinguish between slip happening and stable contact, we take advantage of the intuitive human sense of slip since there are no other more reliable artificial slip sensing solutions available except for success/failure labelling of grasp as been implemented in \cite{li2018slip}, \cite{calandra2017feeling}, which is derivative results of slip or stable contact. For data collection, 12 daily objects are used as shown in Fig. 5. Objects comes in different size, shape and material. During every collection, an object is pressed against sensor surface by human hand, then force are applied to objects to generate target motion in a certain period. The labelling of a specific data is determined by human feeling of slip, translational and rotational slip are included and incipient slip is also labelled into slip category. Every raw data contains 15 frames of $30\times30$ pixel interpolated deformation field images (3-channels: dx, dy and magnitude) with an acquisition rate of 15 HZ. The sensor is fixed on a table during data collection to avoid false labelling. Fig. 6 shows an example of data collection with USB flash disk being the contact object.

In our work, a relative small size of dataset is first manually collected and further slicing to increase dataset volume is performed. Total amount of 1600 data with 800 slip class and 800 non-slip data are obtained. By slicing out subsets from raw data as in \cite{li2018slip}, we increase the dataset size by 5 times to 8000. Five subsets with each containing 10 successive frames are selected out of raw data using a sliding window with stride of 1 frame. As for dataset splitting, 90\% of data is used for training and 10\% for testing. 

\begin{figure}
	\centering
	\includegraphics[width=0.45\textwidth, height=3cm]{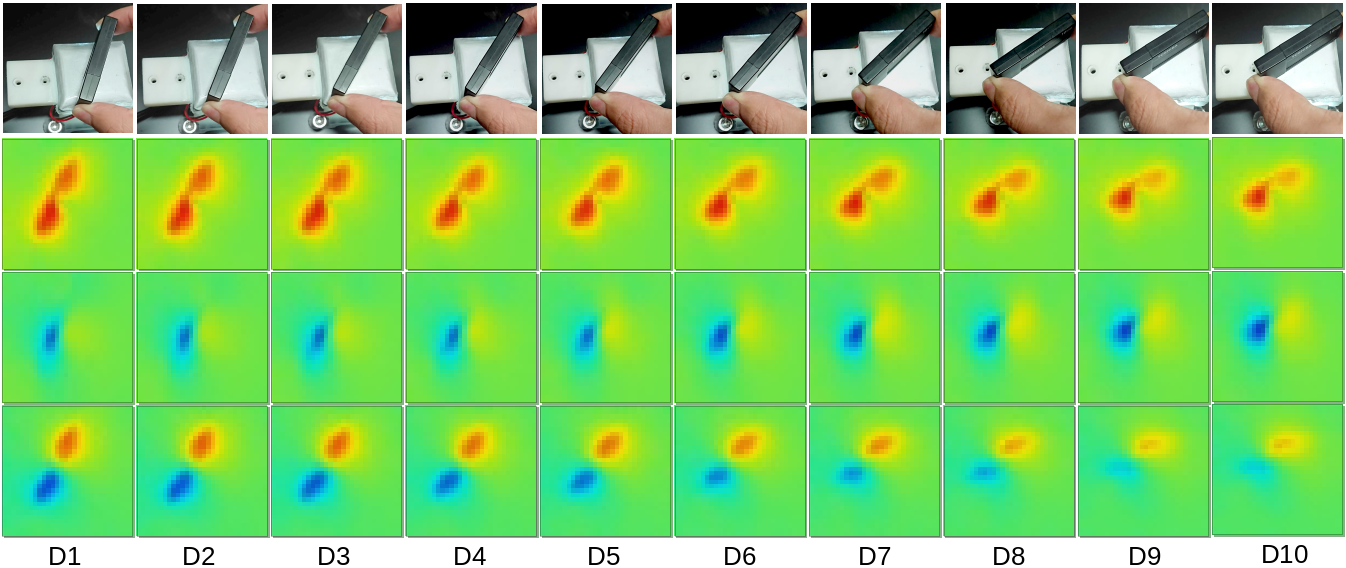}
	\caption{An example of data collection process with USB flash disk. The first row are images recorded by another external camera monitoring the whole process (notes that images in first row are not accurately synchronized to images in later rows). The last 3 rows are magnitude, x axis projection, y axis projection of deformation field respectively.}
\end{figure}

\begin{figure}
	\centering
	\includegraphics[width=0.45\textwidth, height=4cm]{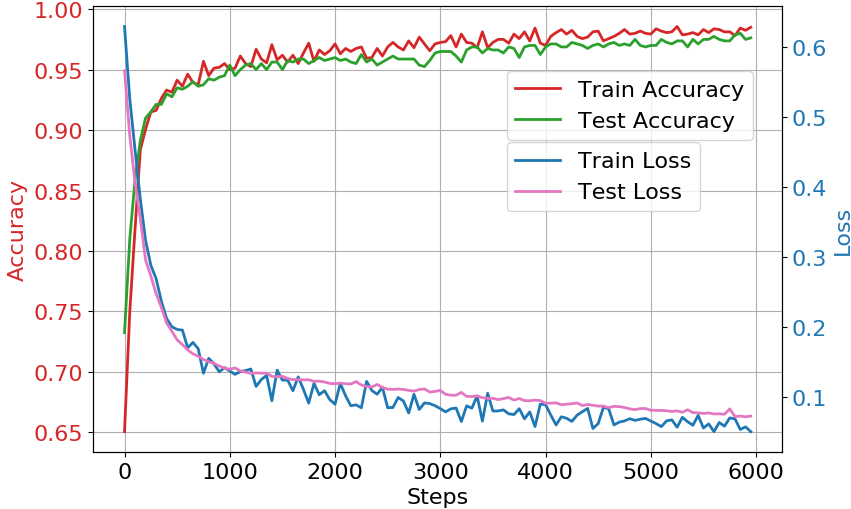}
	\caption{The network model training process.}
\end{figure}

\textbf{Results analysis}. From training history plotted in Fig. 7, we can see that our prediction accuracy on testing dataset rises over 95\% in first 1000 steps with batch size of 32, equivalent to a training time of 6m23s under our hardware setting. \cite{li2018slip} and \cite{van2018slip} are recent works similar to our paper, Table 1 presents a comparison between the performance of our method and these works in terms of accuracy. According to the testing results, our model using sequential frames of interpolated contact deformation field shows higher accuracy than other works. Furthermore, with a much lower input resolution and relatively shallower network structure, our forward pass time is as low as 0.014 seconds, making it a suitable framework to satisfy real-time requirements of robotic manipulation tasks with mobile computing devices. On another perspective, the performance also shows the effectiveness of our tactile sensor on tracking of contact events. For robot grasping problem, our sensor with proper algorithms can surely improve robot performance by providing dynamic spatiotemperal awareness of contact events.

\begin{table}
	\centering
	\caption{Slip detection accuracy of different works. Highest accuracy in each experiments are used.}
	\begin{tabular}{|c|c|c|}
		\hline 
		\textbf{Input} & \textbf{Framework} & \textbf{Accuracy} \\ 
		\hline 
		Interpolated deformation field (image) & ConvLSTM & \textbf{97.62\%} \\
		\hline 
		Tactile and external image \cite{li2018slip} & CNN+LSTM & 88.03\% \\
		\hline 
		Spectral univariate signals \cite{van2018slip} & LSTM & 94.5\% \\ 
		\hline 
		 
	\end{tabular}

\end{table}

\section{Conclusion}

In this work, we develop a optical-based tactile sensor FingerVision. This sensor uses camera as transducing interface and embedded markers as physical feature to track within contact surface. Design, fabrication process and contact deformation extraction algorithms are described in detail. our sensor features stable and high performance in varying environments, multi-axial contact wrench extraction capability, ease of fabrication and readiness to use by various robotic manipulation tasks. Such a sensor is expected to be useful in applications including force feedback grasping, contact area awared in-hand manipulation and slip detection, etc. 

We also propose a framework based on convolutional LSTM network taking deformation field from FingerVision as input to detect contact slip. Our data are sequential image frames and every data is labelled with the aid of human tactile sensing into slip or non-slip categories. The model shows a superior accuracy compared to other sensors and methods. This result not only points out that convolutional LSTM method can be effective on recognizing contact events on optical-based tactile sensor signals, but also takes a glance on how much potential FingerVision sensor has when applied to robot manipulation problems that is hard for existing tactile sensors. 

There remains future works on contact wrench interpretation, and utilization of sensor in real grasping actions. Decoupling of normal, tangential and torsional forces from deformation field to independent values could be very important for problems like force closure estimation \cite{bicchi1995closure}, however this is difficult for the high dimension of sensor output. A possible solution is using deep neural networks to represent the mapping between sensor deformation vector space and wrench space.  Apart from our implementation of slip detection, application to contact force sensing, contact area prediction would be worth investigating.

\section*{Acknowledgment}
Here we would like to thank Lei Tai for his insight into our framework.

\bibliographystyle{IEEEtran}
\bibliography{reference}

\end{document}